# Genetic Algorithm Based Floor Planning System

Hamide Ozlem Dalgic,  Erkan Bostanci, Mehmet Serdar Guzel
SAAT Laboratory, Computer Engineering Department
Ankara University, Golbasi Campus,
Ankara, Turkey
ddalgicozlemm@gmail.com, {ebostanci, mguzel}@ankara.edu.tr

*Abstract*— Genetic Algorithms are widely used in many different optimization problems including layout design. The layout of the shelves play an important role in the total sales metrics for superstores since this affects the customers' shopping behaviour. This paper employed a genetic algorithm based approach to design shelf layout of superstores. The layout design problem was tackled by using a novel chromosome representation which takes many different parameters to prevent dead-ends and improve shelf visibility into consideration. Results show that the approach can produce reasonably good layout designs in very short amounts of time.

*Keywords- market shelves layout, genetic algorithm, DEAP (evolutionary algorithm framework), python,*

## I. Introduction

Genetic Algorithm (GA) is one type of evolutionary algorithms which uses the evolutionary principles found in nature to find the optimal solution of problems [1-2]. Genetic Algorithms work according to probability rules and aim to find an answer for the existing problem and this answer is optimised over generations.

Although the design process for market shelves (Figure 1) seems to be easy, it is an important factor in the revenue and cost.  For example, in a market design, a customer should be able to see all products on the shelves. The market shelves should not occlude each other. Another important factor includes the placement of entrance and exit doors. Customers should be able to navigate easily between shelves until they reach the exit door after entering through the entrance door.

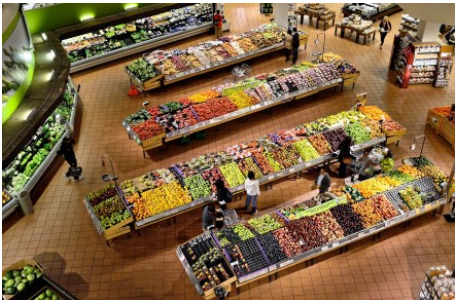

Figure 1. A market shelf layout example

This paper describes the implementation of GA to the Floor Planning System. We describe our approach to designing a chromosome structure and fitness function to yield the optimal layout of the shelves. The result is a 2D map describing the shelf organization.

The rest of the paper is structured as follows: Section II describes the previous studies found in the literature which is followed by a brief explanation of the library used for the evolutionary system, namely Evolutionary Algorithm Framework (DEAP), in Section III. Section IV is dedicated to the description of the chromosome design and other implementation issues. Experimental results are described in Section V. Finally, the paper is concluded and future directions are outlined in Section VI.

## II. Background

Arrangement problems have a very important place when real life problems are considered. Since these problems are difficult to solve, heuristic algorithms such as genetic algorithm are used to produce near optimal results instead of traditional optimization algorithms.

In 2010, facility's layout plan was generated on Two Dimensional Shape Allocation with Genetic Algorithm [3]. Therein, the important point was shapes allocation with minimum space requirement on a 2D surface. Their results showed that the genetic algorithm gives optimal results in two dimensional shape fitting problems [3-4]. Our approach differs from this study in that we mainly focus on navigable corridors for customer walk instead of obtain the minimum space between shapes.

As a similar study, Rojas and Torres [5] designed a system using genetic algorithm to solving bank offices' layouts problem. They represented their problem solution in a chromosome that has five parts. First four parts of the chromosome represent the flanks of the office and fifth part contains the numbers of the departments located in the back part of office.

Maze generation problems are also quite related to floor planning problems [6-7], especially for computer games. Such as automated maze generation for Ms. Pac-Man using Genetic Algorithms. In that study, a genetic algorithm was designed optimal mazes by specifying fitness function to create different mazes which can allow the game to be finished by the player [8]. The navigable corridor issue tackled here is very similar to the one in [8].

## III. DEAP (EVOLUTIONARY ALGORITHM FRAMEWORK)

DEAP is an evolutionary computation algorithm developed by François-Michel De Rainville, Félix-Antoine Fortin and Marc-André Gardner in 2009 with the help of the Python program language [9]. Rapid prototyping allows users to quickly apply their own algorithms instead of limiting them [10]. Therefore, it is fairly easy to implement. The DEAP core consists of two basic structures; creator and toolbox. The Creator module allows generation of genes and populations from data structures such as list, set, dictionary. Toolbox is a structure that contains evolutionary algorithm operators.

## IV. GENETIC ALGORITHM

Genetic Algorithms provide heuristic solutions to complex and difficult problems. GA based on natural biological evolution rules [11-13]. The fundamental principle of the GA is the survival of the fittest as it is in nature. Through the evolutionary process, each generation comes out to have superior features than their parents. To produce a fitter generation in time, the chances of reproduction of failed individuals are reduced. This is provided by several genetic algorithm operators. The most commonly used and important one is crossover operation in which an exchange of sections of the parents' chromosomes is performed.

Yet another GA operation is mutation. Only crossover on chromosomes can be insufficient for variation. Mutation operation ensures that a random modification of the chromosome. Selection of new individuals are made by a fitness function which constitutes the basis of an individual's quality. This function is defined in a problem specific fashion. It assigns a value to all individuals in a population. Selection operation is performed according to these fitness values assigned by fitness function. Outline of a GA can be described with the steps below:

- Creating initial random population.
- Calculating fitness of the each individuals in the population.
- Selection best fit from current population and generate offspring.
- Crossover and mutation operations to provide variety.
- Evaluating fitness of each offspring.
- Replacing failed individuals with newly generated ones.
- Repeating from step two until termination condition reached.

### A. Chromosomal Representation

A solution to our market layout problem is created using blocks to represent market shelves. These blocks are represented in a chromosome which are divided four parts. The first part of the chromosome represents a block's direction; true for vertical blocks and false for horizontal blocks. The second and third parts of the chromosome represents position of block. The last one represents length of the block.

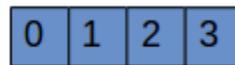

Figure 2. Block representation

```
isVertical = block[0]
xValue = block[1]
yValue = block[2]
blockLength = block[3]
```

An example chromosome may have the values (1,6,7,3) for a block with the following properties: This block was created vertical and 3 units lengths at (6,7) coordinates. This notation made it easier to implement the genetic algorithm and reach the solution.

### B. Initial Population

Initial population is created with certain number of blocks that formed randomly. Each individual is represented by a 10x10 2-dimensional array with 10 blocks. For subsequent graphical display made easy, blocks were represented by 1 and corridors were represented by 0 in the array.

```
population [Individual([[0, 0, 0, 0, 0, 0, 0, 0, 0, 0],
    [0, 0, 0, 0, 0, 0, 0, 0, 0, 0],
    [0, 0, 0, 0, 0, 0, 0, 0, 0, 0],
    [0, 0, 0, 0, 1, 1, 0, 0, 0, 0],
    [0, 0, 0, 0, 1, 1, 0, 1, 0, 0],
    [0, 0, 0, 0, 1, 1, 0, 1, 0, 0],
    [0, 0, 1, 1, 0, 1, 0, 1, 0, 0],
    [0, 0, 0, 0, 0, 0, 0, 0, 0, 0],
    [0, 0, 0, 0, 0, 0, 0, 0, 0, 0],
    [0, 0, 0, 0, 0, 0, 0, 0, 0, 0]])]
```

Figure 3. Map made out from blocks

### C. Fitness Function

Fitness function is defined as a function which determined how good the individuals are as a solution. It assigns a value to each individual in the population considering their suitability for the solution. In GA Based Floor Planning System fitness values are assigned conforming to following purposes; floor design should let go around between all shelves and reaching all products to customers. That navigation control is provided by A* algorithm using a similar approach in [8]. Another criteria for quality of the block is whether blocks are spread homogeneously on the floor and total number of blocks.

According to these constraints, we define the following,

$$r = \begin{bmatrix} 0.05n, n \geq 5 \\ 0.05/n, otherwise \end{bmatrix}$$

$$a = \begin{bmatrix} +0.05, s \\ -0.5, otherwise \end{bmatrix}$$

$$f = b/5.0$$

where
- b = total block count (1's in an individual)
- s = A* algorithm result that the map can complete process
- b = number of neighbours for a block
- r = neighbour count based fitness value
- a = A* algorithm based fitness value
- f = block count based fitness value

$$Fitness = f + a + r$$

*D. Crossover*

The crossover operator is used to create new solutions from existing solutions. This operator exchanges the genes information between individuals and puts together different parts of good solutions for create better solutions [14].

In this problem, crossover operator was applied to blocks instead of individuals. Selected blocks of population individuals were crossed to ensure variety as in Figure 4.

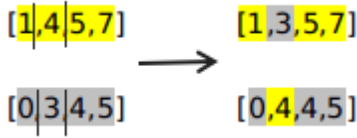

Figure 4. Two point crossover

DEAP's two point crossover method was used for floor planning system. In addition to this there are 2 more crossing methods including one point crossover and uniform crossover methods. In the first stage, a population of 100 individuals were generated. Then this number was increased and the effect of the number of population on the solutions was examined.

*E. Selection*

Selection process determines which individuals are chosen for reproduction. Main principle of the selection is gathering a group of high quality individuals have a higher chance of becoming parents in a mating pool. There are three major types of selection in Genetic Algorithm approach: tournament selection, rank-based selection and roulette wheel selection [15]. In this study, roulette wheel selection was employed.

The probability of selecting an individual in the roulette wheel method is the ratio of the fitness value of that individual to the total fitness values of all individuals as depicted in Figure 5.

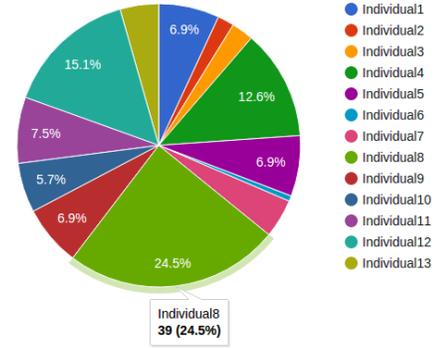

Figure 5. Roulette selection for five candidates with different fitness values.

## V. VISUAL INTERFACE GENERATION

Interface allows easy monitoring the improvements of generations. The python library graphics.py was used to create the interface. Graphics.py is a simple and easy object oriented graphics library written by John Zelle. The library is not a part of standard Python distribution so it must be imported. 500x500 pixels rectangle is created for the floor planning system using graphics.py.

10x10 empty grid is generated in this rectangle. In each individual which is result of crossover 1 values are replaced with a simple 2D box. A map can be directly visualised in this way as shown in Figures 6 and 7.

| 0 | 0 | 0 | 0 | 0 | 0 | 0 | 0 | 0 | 0 |
|---|---|---|---|---|---|---|---|---|---|
| 0 | 0 | 0 | 1 | 0 | 0 | 0 | 0 | 0 | 0 |
| 0 | 1 | 0 | 1 | 0 | 0 | 0 | 0 | 0 | 0 |
| 0 | 1 | 0 | 1 | 0 | 0 | 0 | 0 | 0 | 0 |
| 0 | 1 | 0 | 0 | 0 | 0 | 0 | 0 | 0 | 0 |
| 0 | 1 | 0 | 0 | 0 | 0 | 0 | 0 | 0 | 0 |
| 0 | 1 | 0 | 0 | 0 | 0 | 0 | 0 | 0 | 0 |
| 0 | 0 | 0 | 0 | 1 | 1 | 1 | 0 | 0 | 0 |
| 0 | 0 | 0 | 0 | 0 | 0 | 0 | 1 | 1 | 0 |
| 0 | 0 | 0 | 0 | 0 | 0 | 0 | 0 | 0 | 0 |

Figure 6. Blocks represented by 1s and 0s

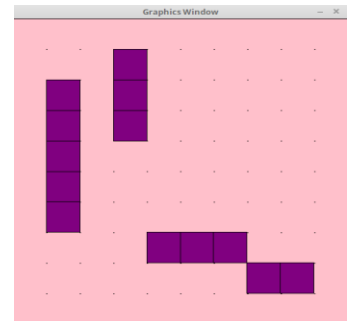

Figure 7. Graphical representation of blocks in Figure 6

## VI. Experimental Results

The result of the study, individuals close to the expected solution are obtained. Low quality individuals are eliminated from the population. It is obtained that fitness values for next generations increased or remained stable in time.

Figure 8 shows how the fitness values change over time for 100 generations.

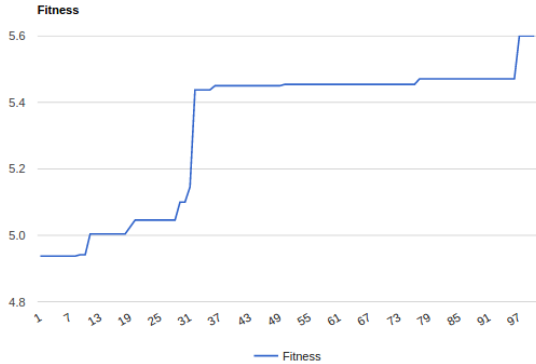

Figure 8. Fitness values alteration of individuals in time

Initial solutions are not ideal since they have many overlapping parts, see Figures 9. Note that these initial layouts are clustered together and hard to navigate.

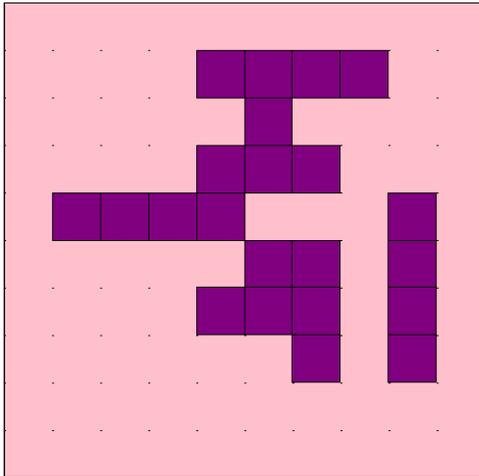

Figure 9. First generation resulting from GA

The algorithm produces better layout designs over time. The evolutionary process eliminates the solutions where navigation is impossible and many overlapping parts exist. Further, the algorithm tries to make the best use of the whole floor by spreading the shelves across the whole surface as shown in the intermediate generation of Figure 11.

Figure 12 and 13 depict the final solutions, two best are shown here. These solutions definitely have navigable corridors where all shelves can be accessed. In addition, the use of the floor is more efficient than the initial solutions.

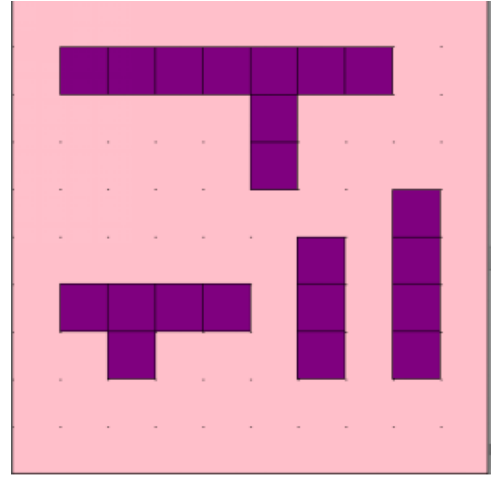

Figure 11. 50 th generation resulting from GA

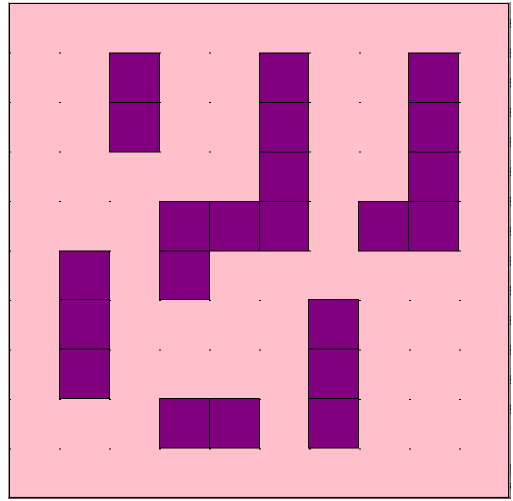

Figure 12. 100 th generation resulting from GA

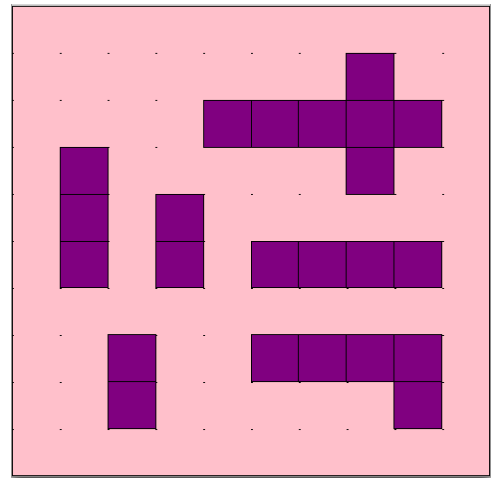

Figure 13. 100 th generation resulting from GA

## VII. CONCLUSION

The genetic Algorithm solution has become an effective algorithm that can be used to achieve fast and easy solution of many difficult and complex problems. This paper provided a solution to the design of market layouts. The ability of the genetic algorithm to obtain more than one optimum solution, possibilities to work with many parameters has also played a significant role in its acceptance. It is observed that the performance of the GA result changes according to given parameters and probability values. Results show that this approach was successful in generating good designs in an automated fashion.

In future works, fitness function will be improved to create better market layout. A way to achieve this is to integrate the shopping basket analysis along with this approach so that the market layout is in an agreement with users' shopping behaviour [16-17].

## AUTHORS' BACKGROUND

| Your Name | Title* | Research Field | Personal website |
| --- | --- | --- | --- |
| Hamide Ozlem Dalgic | Ms. | Computer Engineering | |
| Erkan Bostanci | Asst. Prof. Dr. | Computer Engineering | http://cv.ankara.edu.tr/ebostanci@ankara.edu.tr |
| Mehmet Serdar Guzel | Asst. Prof. Dr. | Computer Engineering | http://cv.ankara.edu.tr/mguzel@ankara.edu.tr |